# On an Ethical Use of Neural Networks:
# A Case Study on a North Indian Raga

**Ripunjai Kumar Shukla and Soubhik Chakraborty**
**Department of Applied Mathematics, BIT Mesra,**
**Ranchi-835215, India**

**ABSTRACT:** The paper gives an artificial neural network (ANN) approach to time series modeling, the data being instance versus notes (characterized by pitch) depicting the structure of a North Indian raga, namely, *Bageshree*. Respecting the sentiments of the artists' community, the paper argues why it is more ethical to model a structure than try and "manufacture" an artist by training the neural network to copy performances of artists. Indian Classical Music centers on the ragas, where emotion and devotion are both important and neither can be substituted by such "calculated artistry" which the ANN generated copies are ultimately up to.

**KEYWORDS:** artificial neural network, raga, time series modeling, ethics in art.

## 1. Introduction

## 1.1 Artificial Neural Network approach to time series modeling

Time series is a series of observations in chronological order, where we clarify that "time" may not mean the time of clock and can well stand for the instance at which an observation is realized. When the linear restriction of the model form is relaxed, the possible number of nonlinear structures that can be used to describe and forecasting a time series is enormous. A good nonlinear model should be "general enough to capture some of the nonlinear phenomena in the data" ([GK92]). Artificial neural networks are one of such models that are able to approximate various nonlinearities in the data. ANNs are flexible computing frameworks for modeling a broad range of nonlinear problems. One significant advantage of the ANN models over other classes of nonlinear model is that ANN's are universal approximators and easily can approximate a large class of functions with a high degree of accuracy. Their power comes from the parallel processing of the information from the data. No prior assumption of the model form is required in the model building process. Instead, the network model is largely determined by the characteristics of the data.

Single hidden layer feed forward network is the most widely used model form for time series modeling and forecasting ([ZPH98]). The model is characterized by a network of three layers of simple processing units connected by acyclic links. The relationship between the output ($y_t$) and the inputs ($y_{t-1}$; $y_{t-2}$; : ; $y_{t-p}$) has the following mathematical representation:

$$y_t = w_0 + \sum_{j=1}^{q} w_j \cdot g\left(w_{0,j} + \sum_{i=1}^{p} w_{i,j} \cdot y_{t-j}\right) + \varepsilon_t \qquad (1)$$

where $w_{i,j}$ (i = 0,1,2,….,p; j = 1,2,…,q) $w_j$ (j = 0,1,2,…,q) and are the model parameters often called the connection weights; p is the number of input nodes and q is he number of hidden nodes. Activation function can take several forms. The type of activation function is indicated by the situation of neuron within the network. In the majority of cases input layer neurons do not have an activation function, as their role is to transfer the input to the hidden layer. The most widely used activation function for the output layer is the linear function as non-linear activation function may introduce distortion to the predicted output. The logistic and hyperbolic functions are often used as the hidden layer transfer functions that are shown in (a) and (b) of equation (2), respectively.

(a)  $\text{Sig}(x) = \dfrac{1}{1 + \exp(x)}$

$$\qquad (2)$$

(b)  $\text{Tanh}(x) = \dfrac{1 - \exp(-2x)}{1 + \exp(-2x)}$

Hence, the ANN model of (1) in fact performs a nonlinear functional mapping from the past observations ($y_{t-1}$; $y_{t-2}$; : : : ; $y_{t-p}$) to the future value $y_t$, i.e.,

$$y_t = f(y_{t-1,\ldots\ldots}y_{t-p},w) + \varepsilon_t$$

where w is a vector of all parameters and f(.) is a function determined by the network structure and connection weights. Thus, the neural network is equivalent to a nonlinear autoregressive model. The simple network given by (1) is surprisingly powerful in that it is able to approximate arbitrary function as the number of hidden nodes q is sufficiently large ([HSW90])**.** In practice, simple network structure that has a small number of hidden nodes often works well in out-of-sample forecasting. This may be due to the over fitting effect typically found in the neural network modeling process. An over fitted model has good fit to the sample used for model building but has poor generalization of the data ([DB04]).

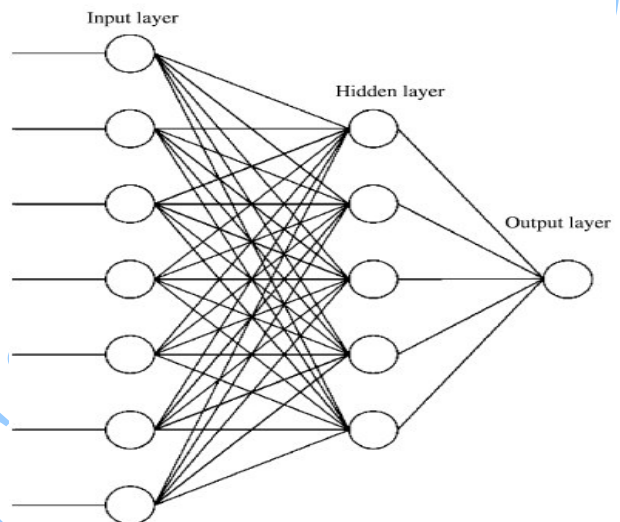

The choice of q is data dependent and there is no systematic rule in deciding this parameter. In addition to choosing an appropriate number of hidden nodes, another important task of ANN modeling of a time series is the selection of the number of lagged observation, p, and dimension of the input vector. This is perhaps the most important parameter to be estimated in ANN model because it plays a major role in determining the non-linear autocorrelation structure of the time series.

There are many different approaches which exist in order to find the optimal architecture of an ANN; theses methods are usually quite complex in nature and are difficult to implement ([ZPH98]). Furthermore, none of these methods can guarantee the optimal solution for all real forecasting problems. There is no simple clear-cut method for determination of these parameters and the usual procedure is to test numerous networks with varying numbers of input and hidden units (p, q), estimate generalization error for each and select the networks with the lowest generalization error ([HLR06]). Once a network structure (p, q) is specified, the network is ready for training a process of parameter estimation. The parameters are estimated such that the cost function of neural network is minimized. Cost function is an overall accuracy criterion such as the following mean squared error:

$$E = \frac{1}{N}\sum_{n=1}^{N}(e_i)^2 = \frac{1}{N}\sum_{n=1}^{N}\left(y_t - \left(w_0 + \sum_{j=1}^{q}w_j \cdot g\left(w_{0j} + \sum_{i=1}^{p}w_{i,j} \ y_{t-1}\right)\right)\right)^2 \quad (3)$$

where, n is the number of error terms. This minimization is done with some efficient nonlinear optimization algorithms other than the basic back propagation training algorithm, in which the parameters of the neural network, $w_{i,j}$, are changed by an amount

$$\Delta w_{i,j} = -\eta \frac{\partial E}{\partial w_{i,j}} \quad (4)$$

where, the parameter η is the learning rate and $\frac{\partial E}{\partial w_{i,j}}$ is the partial derivatives of the function E with respect to the weight $w_{i,j}$. This derivative is commonly computed in two passes. In the forward pass, an input vector from the training set is applied to the input unit of the network and propagated through the network, layer by layer, producing the final output. During the backward pass, the output of the network is compared with the desired output and the resulting error is then propagated through the network, adjusting the weight accordingly. To speed up the learning process, while avoiding the instability of the algorithm, [RM86] introduced a momentum term δ in the Eq. (5), thus obtaining the following learning rule:

$$\Delta w_{i,j}(t+1) = -\eta \frac{\partial E}{\partial w_{i,j}} + \delta \Delta w_{i,j}(t) \quad (5)$$

The momentum term may also be helpful to prevent the learning process from being trapped into the poor local minima, and is usually chosen in the interval [0, 1]. Finally, the estimated model is evaluated using a separate hold-out sample that is not exposed to the training process.

## 1.2 Neural networks and music

The disadvantage of the traditional von Newmann algorithm-based concept is that the processing rules must be clearly specified. In art, especially when it is extempore such as Indian classical music, a better option is to use neural networks where instead of following the traditional pre-programmed layout (which would mean we must specify how to create a certain artwork; in most cases we may not be knowing this "how part" or cannot write it explicitly even if we know), here we would train the network examples of what we want and having the network produce new examples maintaining the style as "learned" from the examples. We refer to [TL91] for an extensive literature on the use of neural networks in music. In particular, we acknowledge the works of Sano, Jenkins, Bharucha, Desain, Honing, Gjerdingen, Todd himself and Mozer to name a few and add the recent use of Hybrid Neural Network (HNN) in artificial composition in the context of Indian classical music by Sinha ([Sin08]). A recent book on algorithmic composition reviewed by the second author is that of Nierhaus ([Nie08]) which rightfully devotes a full chapter to neural network application in art .We quote a paragraph from this review ([Cha09]) that is relevant to our context:

"I have read a paper by Sinha, an independent scholar from India, that attempts an artificial composition in rāgas, using hybrid neural networks (HNNs) ([Sin08]). The attempt deserves a mention here, given that emotion and devotion are of paramount importance in any rāga rendition and neither can be produced by calculated artistry. My personal feeling is that algorithmic composition is more suited to music with fixed scores, such as Western music, than extempore music, such as Indian classical music. However, with or without neural networks, one can certainly simulate a rāga sequence, for instance, by simply exploiting the transition probability matrix of the rāga, computed by an algorithm that uses as input database a very long fixed rāga sequence. This means that the computer is only telling what to play; how to play this simulated rāga sequence--the onsets, duration, loudness of the notes, and, more intricately, the transitory and nontransitory pitch movements between the notes, which capture the emotion--must eventually be up to the composer. Such a composition can be called semi-natural, to differentiate it from a purely artificial composition, where the algorithm decides both what and

45

how to play. A semi-natural composition does make sense, for the computer can simulate randomness more efficiently than a human brain--in fact, "the mathematician Émile Borel claimed the human mind is not able to simulate randomness" ([Mar69]). Any intelligent composer can capitalize on this, especially when composing raga-based songs. Sinha's final comment, "automation should be used as a means of rationalizing human activity rather than a substitute for it" ([Sin08]), supports this view. Perhaps, Nierhaus should continue the debate, by adding a chapter on Indian music (or other Eastern music genres) in the next edition of the book".

A raga, now that it has entered the discussion, should be formally defined. [Cha09] have defined a raga as *a melodic structure with fixed notes and a set of rules characterizing a certain mood conveyed by performance*. For further literature on ragas, see [Jai71]. The Ph.D. thesis of M. R. Gautam, a renowned Indian classical vocalist and musicologist, now available in the form of a book, is also referred to ([Gau08]). For the raga "mood" and its acoustical perspective, the book by Subarnalata Rao ([Rao00]) is a good source of reference. Readers knowing western music but new to Indian music are referred to C. S. Jones ([Jon**]).

Neural network applications in music can be broadly divided into "input" and "output" ([TL91]). Input applications relate to using neural networks for recognition and understanding of a provided stimulus. Our present work comes under this category where a sequence of notes depicting the structure of raga *Bageshree* taken from a standard text ([Dut06]) is used for modeling a time series data; "time" here refers to the instance or serial number of a note and the note itself is characterized by its pitch represented by numbers as given in the Table 1A. In contrast, Sinha's work ([Sin08]) is on music composition and falls in the "output" category. The entire set of *Bageshree* note sequence comprising of 240 notes is given in Table 1B.

**Table 1A: Notes in different octaves and numbers characterizing their pitch**

| C | Db | D | Eb | E | F | F# | G | Ab | A | Bb | B |
|---|---|---|---|---|---|---|---|---|---|---|---|
| **Lower octave** | | | | | | | | | | | |
| *S* | *r* | *R* | *g* | *G* | *M* | *m* | *P* | *d* | *D* | *n* | *N* |
| -12 | -11 | -10 | -9 | -8 | -7 | -6 | -5 | -4 | -3 | -2 | -1 |
| **Middle octave** | | | | | | | | | | | |
| S | r | R | g | G | M | m | P | d | D | n | N |
| 0 | 1 | 2 | 3 | 4 | 5 | 6 | 7 | 8 | 9 | 10 | 11 |
| **S** | **r** | **R** | **g** | **G** | **M** | **m** | **P** | **d** | **D** | **n** | **N** |
| 12 | 13 | 14 | 15 | 16 | 17 | 18 | 19 | 20 | 21 | 22 | 23 |

**Table 1B:** *Bageshree* **Note sequence ([Dut06])**

| Sr.No. | Note | Sr.No. | Note | Sr.No. | Note | Sr.No. | Note | Sr.No. | Note | Sr.No. | Note |
|---|---|---|---|---|---|---|---|---|---|---|---|
| 1 | 0 | 42 | -3 | 83 | 0 | 124 | 10 | 165 | 10 | 206 | 12 |
| 2 | -2 | 43 | -7 | 84 | 0 | 125 | 9 | 166 | 12 | 207 | 10 |
| 3 | -3 | 44 | -2 | 85 | 2 | 126 | 5 | 167 | 14 | 208 | 9 |
| 4 | -2 | 45 | -3 | 86 | 0 | 127 | 9 | 168 | 12 | 209 | 12 |
| 5 | 0 | 46 | -7 | 87 | -2 | 128 | 10 | 169 | 17 | 210 | 10 |
| 6 | 5 | 47 | -3 | 88 | -3 | 129 | 9 | 170 | 15 | 211 | 9 |
| 7 | 5 | 48 | -2 | 89 | -7 | 130 | 3 | 171 | 14 | 212 | 5 |
| 8 | 3 | 49 | 0 | 90 | -3 | 131 | 3 | 172 | 12 | 213 | 9 |
| 9 | 5 | 50 | 5 | 91 | -2 | 132 | 5 | 173 | 17 | 214 | 10 |
| 10 | 9 | 51 | 3 | 92 | -3 | 133 | 5 | 174 | 15 | 215 | 9 |
| 11 | 10 | 52 | 2 | 93 | 0 | 134 | 9 | 175 | 14 | 216 | 12 |
| 12 | 9 | 53 | 0 | 94 | 5 | 135 | 9 | 176 | 12 | 217 | 9 |
| 13 | 5 | 54 | 2 | 95 | 3 | 136 | 10 | 177 | 10 | 218 | 10 |
| 14 | 10 | 55 | 0 | 96 | 5 | 137 | 9 | 178 | 12 | 219 | 12 |
| 15 | 9 | 56 | -2 | 97 | 9 | 138 | 12 | 179 | 14 | 220 | 9 |
| 16 | 5 | 57 | -3 | 98 | 5 | 139 | 9 | 180 | 10 | 221 | 10 |
| 17 | 9 | 58 | -7 | 99 | 9 | 140 | 10 | 181 | 12 | 222 | 9 |
| 18 | 10 | 59 | -3 | 100 | 9 | 141 | 12 | 182 | 10 | 223 | 12 |
| 19 | 12 | 60 | 0 | 101 | 10 | 142 | 9 | 183 | 9 | 224 | 10 |
| 20 | 12 | 61 | 0 | 102 | 9 | 143 | 10 | 184 | 14 | 225 | 12 |
| 21 | 10 | 62 | -2 | 103 | 5 | 144 | 9 | 185 | 10 | 226 | 14 |
| 22 | 9 | 63 | -3 | 104 | 7 | 145 | 12 | 186 | 9 | 227 | 10 |
| 23 | 5 | 64 | 0 | 105 | 9 | 146 | 10 | 187 | 5 | 228 | 12 |
| 24 | 7 | 65 | -2 | 106 | 3 | 147 | 9 | 188 | 9 | 229 | 10 |
| 25 | 9 | 66 | 0 | 107 | 5 | 148 | 5 | 189 | 5 | 230 | 9 |
| 26 | 5 | 67 | 5 | 108 | 3 | 149 | 9 | 190 | 10 | 231 | 5 |
| 27 | 3 | 68 | 3 | 109 | 2 | 150 | 10 | 191 | 10 | 232 | 7 |
| 28 | 5 | 69 | 5 | 110 | 0 | 151 | 12 | 192 | 9 | 233 | 9 |
| 29 | 3 | 70 | 9 | 111 | 0 | 152 | 10 | 193 | 5 | 234 | 5 |
| 30 | 2 | 71 | 10 | 112 | -2 | 153 | 9 | 194 | 9 | 235 | 3 |
| 31 | 0 | 72 | 9 | 113 | 0 | 154 | 5 | 195 | 12 | 236 | 2 |
| 32 | -3 | 73 | 5 | 114 | 5 | 155 | 7 | 196 | 12 | 237 | 0 |
| 33 | -2 | 74 | 10 | 115 | 3 | 156 | 9 | 197 | 17 | 238 | -2 |
| 34 | 0 | 75 | 9 | 116 | 5 | 157 | 5 | 198 | 15 | 239 | -3 |
| 35 | 5 | 76 | 5 | 117 | 10 | 158 | 3 | 199 | 17 | 240 | 0 |
| 36 | 0 | 77 | 7 | 118 | 9 | 159 | 2 | 200 | 15 | | |
| 37 | -2 | 78 | 9 | 119 | 10 | 160 | 0 | 201 | 14 | | |
| 38 | -3 | 79 | 3 | 120 | 12 | 161 | 5 | 202 | 12 | | |
| 39 | -7 | 80 | 5 | 121 | 14 | 162 | 3 | 203 | 12 | | |
| 40 | -3 | 81 | 3 | 122 | 10 | 163 | 5 | 204 | 14 | | |
| 41 | -2 | 82 | 2 | 123 | 12 | 164 | 9 | 205 | 10 | | |

**Abbreviations:**
The letters S, R, G, M, P, D and N stand for Sa (always Sudh), Sudh Re, Sudh Ga, Sudh Ma, Pa (always Sudh), Sudh Dha and Sudh Ni respectively. The letters r, g, m, d, n represent Komal Re, Komal Ga, Tibra Ma, Komal Dha and Komal Ni respectively. A note in Normal type indicates that it belongs to middle octave; if in italics it is implied that the note belongs to the octave just lower than the middle octave while a bold type indicates it belongs to the octave just higher than the middle octave. Sa is the tonic in Indian music.

*Bageshree* uses the notes S, R, g, M, P, D and n. Before presenting the results and a discussion in the next section, here are some fundamental features of raga Bageshree:

**Raga: *Bageshree* ([Dut06])**
*Thaat* (raga-group according to the permitted notes) : Kafi
*Aroh* (ascent): S g M D n **S**   OR   S *n D n* S M, g M D n **S**
*Awaroh* (descent)**: S** n D, M PD g, M g R S
*Jati*: Aurabh-Sampoorna (jati is just another way of grouping of raga reflecting no. of distinct notes used in ascent-descent; Aurabh-Sampoorna means 5 distinct notes are used in ascent and 7 in descent)
*Vadi Swar* (most important note): M (some say D)
*Samvadi Swar* (second most important note): S (some say g)
*Prakriti* (nature): restful
*Pakad* (catch):  S *n D*, S M D n D, M g R S
*Nyas swars* (Stay notes): g M D
Time of rendition: 9 PM to 12 PM

**Comment:** P is a weak note (*Alpvadi swar*) in this raga. Notes that are important but different from *Vadi* and *Samvadi* are called *Anuvadi*. For example, if one takes D and g as *Vadi-Samvadi*, then S, R, M and n will be Anuvadi. Notes not permitted in a raga are called *Vivadi*. Here r, G, m, d and N are *vivadi*.

Here is Rajan P. Parrikar's description of this raga in his own words[*]:   (http://www.sawf.org/newedit/edit12022002/musicarts.asp;   the single quote refers to lower octave, the double quote higher octave, the braces a "kan" swar(note) where a faint touch is to be given to a note before we move to the next note)

---

[*] Reproduced with permission.

"Let us now examine the structure of Bageshree through a set of characteristic phrases.

**S, n' D', D' n' S M, M g, (S)R, S**
The madhyam is very powerful, the melodic centre of gravity, as it were. In the avarohi mode, a modest pause on **g** is prescribed (it helps prevent a spill of Kafi). Also notice the kaN of **S** imparted to **R.**

**S g M, M g M D, D n D, D-M, g, (S)R, S**
The Raganga germ is embedded in this tonal sentence. Of particular interest are the nyAsa bahutva role of **D** and the **D-M** sangati.

**g M D n S"**
**g M n D n S"**
**M D n S"**
Each of these Arohi strips is a candidate for the uttarAnga launch.

**S", (n)D n D, M P D (M)g, (S)R, S**
This phrase illustrates the modus operandi for insertion of the pancham within a broader avarohi context. The touch of **P** occasions moments of delicious frisson. In some versions, **P** is explicity summoned in Arohi prayogas such as **D n D, P D n D.** When the clips roll out there will be opportunities aplenty to sample a variety of procedures concerning the pancham.

**S g M, M g, RgM, M g (S)R, S**
Observe the vakra Arohi prayoga of **R.** A thoughtless or cavalier approach here can lead to an inadvertent run-in with Abhogi (recall the ucchAraNa bheda that separates Bageshree and Abhogi in this region of the poorvAnga) and run afoul of the Bageshree spirit."

## 2. Results

Various models of ANN family were tried with various input lags to capture the trend of the data series are listed in **Table 1C**. From these fitted models, $N^{2-4-1}$ model is found suitable as it modeled the original data comparatively better than the rest of models. The estimated parameters of model $N^{2-4-1}$ is shown in **Table 2.** Its RMSE (Root Mean Square Error) and MAE (Mean Absolute Error) values were the lowest among all the fitted models (see appendix for the formulae of RMSE and MAE). In **Figure 1** the network architecture is shown. It includes only two inputs ($Y_{t-1}$ and $Y_{t-2}$) with four hidden units in the hidden layer. **Figure 2** shows the graph between the observed and expected values through the ANN model, which maximally occupies the trend of observed data series. In other words, it is quite possible to capture the raga structure through an artificial neural network. The argument applies to a performance as well but the note sequence has to be extracted first using acoustical signal processing.

The ethical issue related to performance is discussed in the next section.

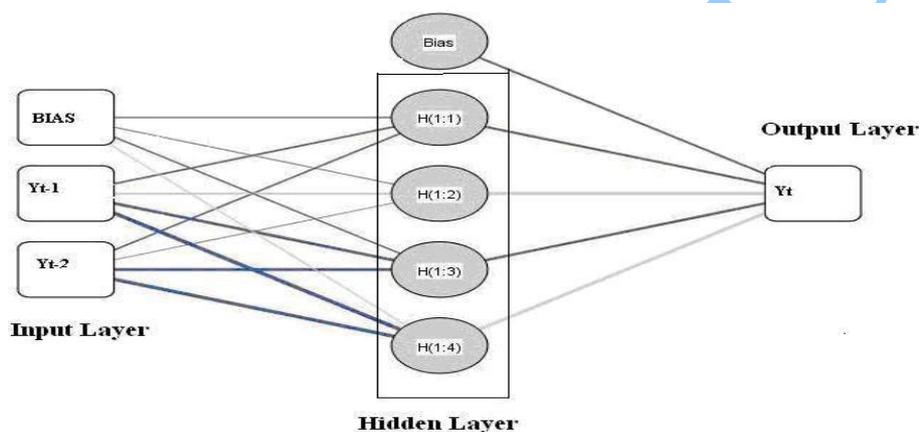

**Figure 1. Network architecture of $N^{2-4-1}$**

## Table 1C: ANN models on Music data

| Sr. No. | ANN Type | No. of Units in Hidden layer | Activation func. for Input | Activation func. for Output | RMSE | MAE |
|---|---|---|---|---|---|---|
| 1 | $N^{1-1-1}$ | 1 | Tanh | Identity | 3.895 | 3.255 |
| 2 | $N^{1-2-1}$ | 2 | Tanh | Tanh | 3.708 | 3.034 |
| 3 | $N^{1-3-1}$ | 3 | Tanh | Tanh | 2.725 | 2.258 |
| 4 | $N^{1-4-1}$ | 4 | Tanh | Tanh | 2.629 | 2.202 |
| 5 | $N^{1-5-1}$ | 5 | Tanh | Tanh | 2.627 | 2.205 |
| 6 | $N^{1-4-1}$ | 4 | Tanh | Identity | 2.917 | 2.407 |
| 7 | $N^{2-4-1}$ | 4 | Tanh | Identity | 2.595 | 2.172 |
| 8 | $N^{2-4-1}$ | 4 | Tanh | Tanh | 3.089 | 2.498 |
| 9 | $N^{2-5-1}$ | 5 | Tanh | Identity | 3.081 | 2.455 |
| **10** | **$N^{2-4-1}$** | **4** | **Sigmoid** | **Identity** | **2.521** | **2.071** |
| 11 | $N^{2-4-1}$ | 5 | Tanh | Sigmoid | 2.643 | 2.191 |
| 12 | $N^{3-1-1}$ | 1 | Tanh | Identity | 2.701 | 2.134 |
| 13 | $N^{3-2-1}$ | 2 | Tanh | Identity | 2.658 | 2.152 |
| 14 | $N^{3-3-1}$ | 3 | Tanh | Identity | 2.561 | 2.137 |
| 15 | $N^{3-4-1}$ | 4 | Tanh | Identity | 2.656 | 2.195 |
| 16 | $N^{3-4-1}$ | 4 | Sigmoid | Identity | 2.699 | 2.171 |
| 17 | $N^{3-4-1}$ | 4 | Tanh | Tanh | 2.654 | 2.165 |
| 18 | $N^{3-4-1}$ | 4 | Tanh | Sigmoid | 2.772 | 2.205 |
| 19 | $N^{3-3-1}$ | 3 | Tanh | Sigmoid | 2.616 | 2.129 |
| 20 | $N^{3-3-1}$ | 3 | Sigmoid | Identity | 2.564 | 2.147 |
| 21 | $N^{4-3-1}$ | 3 | Tanh | Identity | 2.717 | 2.247 |
| 22 | $N^{4-3-1}$ | 3 | Tanh | Tanh | 2.681 | 2.199 |
| 23 | $N^{4-3-1}$ | 3 | Tanh | Sigmoid | 2.787 | 2.243 |
| 24 | $N^{4-4-1}$ | 4 | Sigmoid | Identity | 2.753 | 2.25 |
| 25 | $N^{4-3-1}$ | 3 | Sigmoid | Identity | 2.606 | 2.154 |
| 26 | $N^{4-4-1}$ | 4 | Sigmoid | Identity | 2.626 | 2.158 |
| 27 | $N^{4-5-1}$ | 5 | Sigmoid | Identity | 2.710 | 2.183 |
| 28 | $N^{4-5-1}$ | 5 | Tanh | Sigmoid | 2.789 | 2.249 |
| 29 | $N^{4-6-1}$ | 6 | Sigmoid | Identity | 2.583 | 2.148 |
| 30 | $N^{5-4-1}$ | 4 | Tanh | Identity | 2.943 | 2.421 |
| 31 | $N^{5-4-1}$ | 4 | Sigmoid | Identity | 2.874 | 2.377 |
| 32 | $N^{5-5-1}$ | 5 | Tanh | Identity | 2.767 | 2.173 |
| 33 | $N^{5-4-1}$ | 4 | Tanh | Tanh | 2.917 | 2.331 |
| 34 | $N^{5-6-1}$ | 6 | Tanh | Identity | 2.664 | 2.17 |
| 35 | $N^{5-7-1}$ | 7 | Tanh | Identity | 2.587 | 2.161 |
| 36 | $N^{5-6-1}$ | 6 | Tanh | Sigmoid | 2.713 | 2.154 |
| 37 | $N^{5-5-1}$ | 5 | Tanh | Sigmoid | 2.724 | 2.199 |
| 38 | $N^{5-6-1}$ | 6 | Sigmoid | Tanh | 2.703 | 2.166 |

**Table 2. Estimated parameters of ANN N$^{2-4-1}$ model**

| Parameter Estimates | | | | | | |
|---|---|---|---|---|---|---|
| Predictor | | Predicted | | | | |
| | | Hidden Layer | | | | Output Layer |
| | | H(1:1) | H(1:2) | H(1:3) | H(1:4) | Y$_t$ |
| **Input Layer** | (Bias) | -.621 | -.358 | -.641 | .429 | |
| | Y$_{t-1}$ | -.832 | 1.034 | -1.754 | -3.818 | |
| | Y$_{t-2}$ | -.776 | -.009 | -1.831 | -2.988 | |
| **Hidden Layer** | (Bias) | ---- | ---- | ---- | ---- | -.852 |
| | H(1:1) | ---- | ---- | ---- | ---- | -1.202 |
| | H(1:2) | ---- | ---- | ---- | ---- | 2.845 |
| | H(1:3) | ---- | ---- | ---- | ---- | -1.206 |
| | H(1:4) | ---- | ---- | ---- | ---- | 1.222 |

The MAE (Mean Absolute Error) and RMSE (Root Mean Square Error), which are computed from the following equations, are employed as performance indicators in order to measure forecasting performance of the fitted model in comparison with those other fitted models.

$$MAE = \frac{1}{N}\sum_{i=1}^{N}|ei|$$

$$RMSE = \frac{1}{N}\sum_{i=1}^{N}(ei)2$$

where e$_i$ is the i[th] (i=1,2,…,N) residual of the fitted model. The lower value of MAE and RMSE will indicate the better fit of the model.

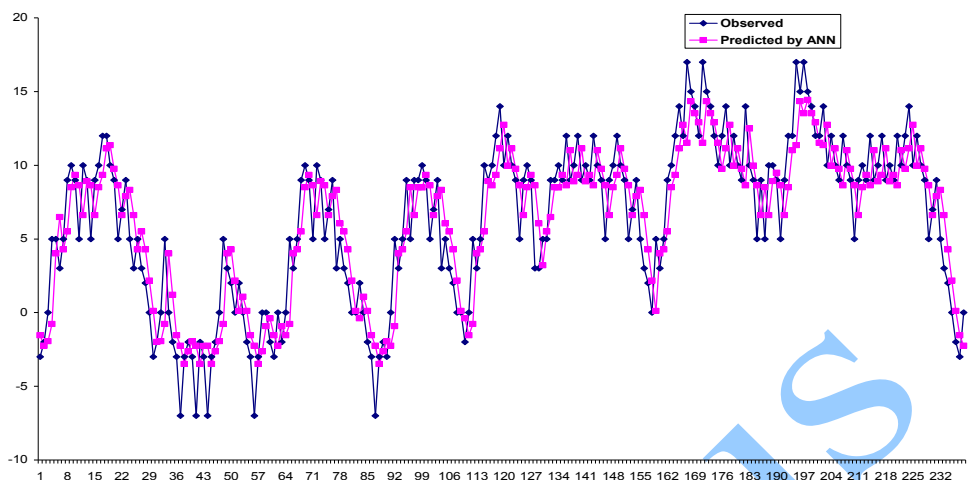

**Figure 2. Observed and Predicted through ANN (N$^{2\text{-}4\text{-}1}$) model**

## 3. Discussion

Perhaps neural networks can "learn" to produce a work of art very close to some celebrated artist, where we further agree, that this closeness can be narrowed by increasing the database training samples significantly. This would undisputedly be a scientific invention, but is this a substitute for the emotion and devotion that the original work of the artist had behind it? This is precisely where the crux of the debate lies. If neural networks assist in learning, e.g., by reducing the time of the learning process- this is generally very long in Indian classical music- it makes some sense. If the motive is to "manufacture" an artist, we strongly object to it as being unethical. In the final analysis, therefore, it is not the invention but the motive of the inventor that comes under scrutiny. Perhaps the reader can now speculate why we chose to model a raga-structure rather than a raga-performance using neural networks! All we wanted is to have the neural network "learn" some general features about raga *Bageshree* as given by its specific note combinations given in the note sequence and then use what is "learnt" to model the raga structure. As observed in section 2, **N$^{2\text{-}4\text{-}1}$** model is found to be appropriate as it modeled the original data comparatively better than the rest of the models under investigation. In this finding, no specific artist was targeted for utilization in the training. Nobody's sentiments are hurt, no ethics broken**.** Is this not one of the right ways of using ANN in an art form that does not lead to ethical conflicts?

**4. Concluding Remarks**

As Sinha ([Sin08]) has put it correctly, neural networks should be used to rationalize human activities rather than as a substitute thereof. Describing a raga structure through neural networks is rational, so is assisting in training of a budding musician. In other words, we would highly appreciate if neural networks can help understand music better. However, any attempt to "manufacture" a *Mozart*, *Beethoven* or *Baba Allauddin Khan* must be opposed. Each of such maestros attained unparalleled musical heights only after years of dedicated hard work, their individual genius notwithstanding. The devotion and perseverance of the artist receive a setback by a computer "manufactured" copy. Moreover, from a psychological point of view, *it is essential to be emotional for being an artist (just as it is essential to be logically sound for being a scientist*) so that all artists are inherently emotional creatures. It then becomes a moral responsibility of the scientific community, which claims to be *the most rational*, to protect the sentiments of these maestros. This should also go a long way in reducing the gap, if any, between musicians and scientists, thereby raising the standard of scientific research in music and other art forms to greater heights. [Concluded]

**Acknowledgement**

[*] It is a pleasure to thank Mr. Rajan P. Parrikar for granting us the permission to quote from his musical archive**.** Parrikar is a recognized expert in Indian classical music.